\documentclass[a4paper]{article}
\usepackage[affil-it]{authblk}
\usepackage{times}
\usepackage{graphicx}
\usepackage{latexsym}
\usepackage{hhline}
\usepackage{amsmath}
\usepackage{amssymb}
\usepackage{multicol}

\usepackage{tikz}
\usepackage{bm}
\usepackage{pgfplots}
\usetikzlibrary{pgfplots.groupplots}
\usetikzlibrary{positioning,shapes,shadows,arrows}

\usetikzlibrary{shapes.multipart}
\usetikzlibrary{automata}
\usetikzlibrary{topaths}
\newcommand \defn {\mathrel{\triangleq}}
\newcommand \argmax{\mathop{\rm arg\,max}}

\tikzstyle{stone}= [circle,ball color = gray!10,inner sep=0mm, minimum size=12mm]
\tikzstyle{wood} = [circle,ball color = brown!60,draw=black,inner sep=0mm, minimum size=12mm]
\tikzstyle{ice}=[circle, ball color = cyan,inner sep=0mm, minimum size=12mm]
\tikzstyle{pig}=[circle, ball color = green,fill=green,inner sep=0mm, minimum size=12mm]

\tikzstyle{attribute} = [top color = white, bottom color = blue!30, draw=blue!50!black!100, drop shadow,text centered, rounded corners, text width = 8cm, minimum height = 1.5cm]

\tikzstyle{arrow} = [single arrow, single arrow head indent = 1pt, rotate = -90, minimum width = 1.5cm, minimum height = 1.5cm,  top color =  blue!30, bottom color = white, draw=blue!50!black!100, drop shadow]
\tikzstyle{thinarrow} = [->,>=latex,thick,line width = 0.1cm, color=blue!50]

\title{A Bayesian Ensemble Regression Framework on the Angry Birds Game}

\author{Nikolaos Tziortziotis, Georgios Papagiannis, Konstantinos Blekas \\ 
Department of Computer Science \& Engineering \\  University of Ioannina, Greece \\
email: \{ntziorzi,gpapagia,kblekas\}@cs.uoi.gr }

\begin{document}

\maketitle

\bibliographystyle{ecai2014}

\begin{abstract}
An ensemble inference mechanism is proposed on the Angry Birds 
domain. It is based on an efficient tree structure for encoding
and representing  game screenshots, where it exploits its enhanced 
modeling capability. 
This has the advantage to establish an informative feature space
and modify the task of game playing to 
a regression analysis problem. To this direction, we assume that
each type of object material and bird pair has its own Bayesian
linear regression model. In this way, a multi-model regression framework
is designed that simultaneously calculates the conditional expectations
of several objects and 
makes a target decision through an ensemble of regression models.
The learning procedure is performed according to an online estimation
strategy for the model parameters.
We provide comparative experimental results on several game levels 
that empirically illustrate the efficiency of the proposed methodology.
\end{abstract}

\section{INTRODUCTION}

Angry birds  was first launched five years ago by Rovio(TM), and since then it has become one of the most popular games nowadays. The objective is to get rid of the pigs, which are usually protected in structures made of different kinds of building materials,  by killing them. This is achieved by taking control of a limited number of various birds' types, which the player launches to the targets (e.g. building blocks or pigs) via a slingshot. It must be noted that different types of birds are available with some of them being more effective against particular materials, while some other have special features as will be discussed later. The received return at each level is calculated according to the number of pigs killed, the number of the unused birds as well as to the destruction on the structure that  achieved. Roughly speaking, the fewer birds are used as well as the more damage to the structures achieved, the higher the received return.

Due to its nature (e.g. large state and action spaces, continuous tap timing, various objects' properties, noisy object detection, inaccurate physical models),  Angry Birds constitute a really challenging task. During the last two years, a number of works have been proposed which are focused on the development of AI agents with playing capabilities similar to those exhibited by human
players. The Angry birds competitions\footnote{https://aibirds.org/} poses several challenges for
building various AI approaches. A basic game platform \cite{Ge14} is provided by the organisers, that makes use of the Chrome version of the Angry Birds  and incorporates a number of components such as, computer vision, trajectory planning, game playing interface which can be freely used for the agent construction.

Two different machine learning techniques, the Weight Majority algorithm and the Naive Bayesian Network,  have been applied in \cite{Chen13} for selecting the most appropriate shot at each time step. However, the depicted feature space is extremely large since it incorporates a large amount of information about the scene of the game. In addition, it requires a preprocessing step over the input data in order to separate them among positive (shots in winning games) and negative (shots in losing games) examples. 
In \cite{Ferreira13,Lin13} a qualitative spatial representation and reasoning framework has been
introduced that is capable of extracting decision rules according to structural properties. Finally, a model based approach has been presented in \cite{Polceanu13} which tries to learn the environmental model. Then, a number of trajectories are tested in the approximated model by performing 
a maximum impact selection mechanism.

In this work, we propose a Bayesian ensemble regression framework for designing an
intelligent agent for the Angry Bird domain. 
The main advantages of our approach lies on two aspects: 
\begin{itemize} 
\item Firstly, a novel tree structure is proposed for mapping scenes
of game levels, where the nodes represent different material of solid objects. 
This state representation is informative as incorporates all the necessary
knowledge about game snapshots, and simultaneously abstract so as to reduce
the computational cost and accelerate the learning procedure. 
This tree representation allows the construction of an efficient and powerful 
feature space that van be used next for the prediction.
\item Secondly, an ensemble learning approach \cite{Moreira12} is designed where every possible 
pair of `object material' - `bird type' has its own Bayesian linear regression 
model for calculating the expected reward. An ensemble integration framework
based on the UCB algorithm \cite{Auer02} is employed using the predictions
to obtain the final ensemble prediction. Then, an online estimation procedure is
performed in order to adjust the regression model parameters. Finally, an appropriate
Gaussian kernel space has been constructed by using a clustering
procedure to a randomly selected data collection. 
\end{itemize}

The remainder of paper is organised as follows. The general framework of our methodology is described in Section 2.
In particular, the proposed tree structure which is the main building block
in our approach, together with the ensemble mechanism of linear regressors
are presented. Furthermore, some issues are discussed about the 
{\em feasibility} property of tree nodes, as well as about the {\em tap timing} procedure. 
To assess the performance of the proposed methodology we present 
in Section 3 numerical experiments on the `Poached Eggs' game set 
and give some initial comparative results with the \emph{naive} agent 
provided by the organisers. 
Finally, in Section 4 we provide conclusions and suggestions for future research.


\section{PROPOSED STRATEGY}

Our work is based on the project Angry Bird Game
Playing software (version 1.31).
The proposed methodology is focused on establishing an efficient 
state space representation, so as to incorporate all the useful
information of objects from Angry Birds levels as recognized by the game vision 
system. In addition, a decision making mechanism has been
designed using an Bayesian ensemble regression framework in order to 
discover the optimum policy and obtain the final ensemble prediction.

\begin{figure}[h!]
  \centering
  \scalebox{0.6}{\begin{tikzpicture}[node distance = 0.0cm]

\node[attribute] (first) at (0,0) {\LARGE{\textbf{1. Tree structure construnction}}};
\node[arrow] (arrow1) at (0,-1.4){};
\node[attribute] (second) at (0,-3.1) {\LARGE{\textbf{2. Feasibility examination}}};
\node[arrow] (arrow2) at (0,-4.5) {};
\node[attribute] (third)  at (0,-6.2) {\LARGE{\textbf{3. Prediction: expected reward calculation}}};
\node[arrow] (arrow3) at (0,-7.6) {};
\node[attribute] (forth) at (0,-9.3) {\LARGE{\textbf{4. Target and tap time selection}}};
\node[arrow] (arrow4) at (0,-10.7) {};
\node[attribute] (fifth) at (0,-12.4) {\LARGE{\textbf{5. Regression model parameters adjustment}}};
\draw[thinarrow] (fifth.south) -- ++(0,-1.0) -- ++(6.0,0) -- ++(0,14.2) -- ++([xshift=2.2cm]first.west);
\end{tikzpicture}}
  \caption{Flow diagram of the proposed method} 
\label{fig:flowdiagram}
\end{figure}
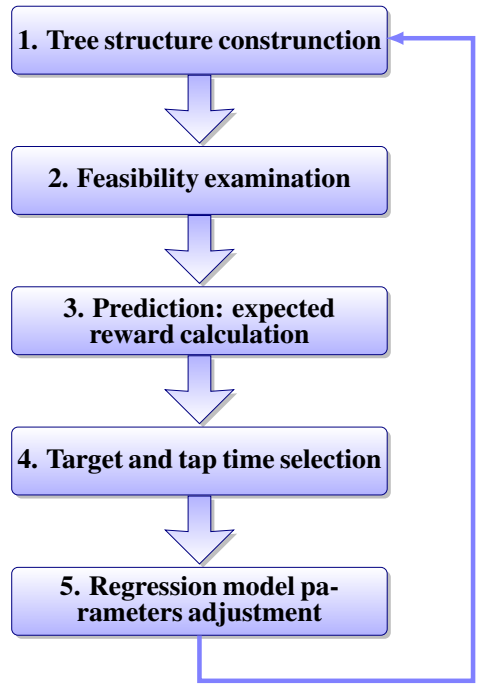

Figure \ref{fig:flowdiagram} illustrates briefly the proposed approach.
A step-by-step description is the following:
\begin{enumerate}
\item Construct the {\bf tree structure} of the game scene and evaluate 
each node.
\item Examine the {\bf feasibility of nodes} in terms of 
their ability to be reached and become possible targets.
\item Calculate the expected reward of each feasible node (target) 
according to a {\bf Bayesian ensemble regression} scheme, which takes into
account the type of object material, as well as the bird. The
optimum target is then selected.
\item Perform shooting according to a {\bf tap timing} procedure.
\item Adjust the model parameters of the selected regressor using 
an online {\bf learning procedure}.
\end{enumerate} 
Next, we give a detailed description of the main building blocks 
of our methodology.

\subsection{An advanced tree-structure for the Angry Birds scene representation}
 
The input in our scheme is the game scene consists of a list of (dynamic or static) 
objects together with some measurements of them, as taken by the Angry Bird vision system.
We have considered seven (7) types of materials for objects presented in the game:
\begin{itemize}
\item Ice/Glass (I)
\item Wood (W)
\item Stone (S)
\item Rolling Stone (RS)
\item Rolling Wood (RW)
\item Pig (P)
\item TNT (T)
\end{itemize}

\begin{figure}[h!]
  \centering
  \scalebox{0.47}{\begin{tikzpicture}
\node[wood] at (0,0) (n11) {\Huge{$\mathbf{s_{11}}$}};
\node[wood] at (2,0) (n12) {\Huge{$\mathbf{s_{12}}$}};
\node[wood] at (4,0) (n13) {\Huge{$\mathbf{s_{13}}$}};
\node[wood] at (6,0) (n14) {\Huge{$\mathbf{s_{14}}$}};
\node[wood] at (8,0) (n15) {\Huge{$\mathbf{s_{15}}$}};
\node[ice] at (3,2) (n21) {\Huge{$\mathbf{s_{21}}$}};
\node[ice] at (5,2) (n22) {\Huge{$\mathbf{s_{22}}$}};
\draw (n13) -- (n21);
\draw (n13) -- (n22);
\node[wood] at (4,4) (n31) {\Huge{$\mathbf{s_{31}}$}};
\draw (n21) -- (n31);
\draw (n22) -- (n31);
\node[wood] at (4,6) (n41) {\Huge{$\mathbf{s_{41}}$}};
\path (n12) edge [out=90,in=240] (n41);
\path (n14) edge [out=90,in=300] (n41);
\draw (n31) -- (n41);
\node[wood] at (2,8) (n51) {\Huge{$\mathbf{s_{51}}$}};
\node[wood] at (4,8) (n52) {\Huge{$\mathbf{s_{52}}$}};
\node[wood] at (6,8) (n53) {\Huge{$\mathbf{s_{53}}$}};
\draw (n41) -- (n51);
\draw (n41) -- (n52);
\draw (n41) -- (n53);
\node[pig] at(4,10) (n61) {\Huge{$\mathbf{s_{61}}$}};
\draw (n52) -- (n61);
\node[wood] at (4,12) (n71) {\Huge{$\mathbf{s_{71}}$}};
\draw (n61) -- (n71);
\path (n51) edge [out=90,in=240] (n71);
\path (n53) edge [out=90,in=300] (n71);
\node[stone] at (4,14) (n81) {\Huge{$\mathbf{s_{81}}$}};
\draw (n71) -- (n81);
\node[wood] at (4,16) (n91) {\Huge{$\mathbf{s_{91}}$}};
\draw (n81) -- (n91);
\node[rectangle, fill=white] at (4,18) (n101) {\Huge{\textbf{Root}}};
\draw[dashed] (n91) -- (n101);
\path[dashed] (n11) edge [out=90,in=240] (n101);
\path[dashed] (n15) edge [out=90,in=300] (n101);
\node at (-2.0,0) (l1) {\huge{\textbf{Level 1}}};
\node at (-2.0,2) (l2) {\huge{\textbf{Level 2}}};
\node at (-2.0,4) (l3) {\huge{\textbf{Level 3}}}; 
\node at (-2.0,6) (l4) {\huge{\textbf{Level 4}}};
\node at (-2.0,8) (l5) {\huge{\textbf{Level 5}}};
\node at (-2.0,10) (l6) {\huge{\textbf{Level 6}}};
\node at (-2.0,12) (l7) {\huge{\textbf{Level 7}}};
\node at (-2.0,14) (l8) {\huge{\textbf{Level 8}}};
\node at (-2.0,16) (l9) {\huge{\textbf{Level 9}}};
\end{tikzpicture}}
  \includegraphics[scale=0.73]{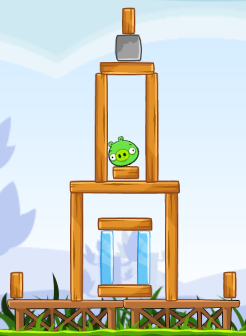}
  \caption{The proposed tree structure consisting of $16$ nodes at the first game level.} 
\label{fig1}
\end{figure}
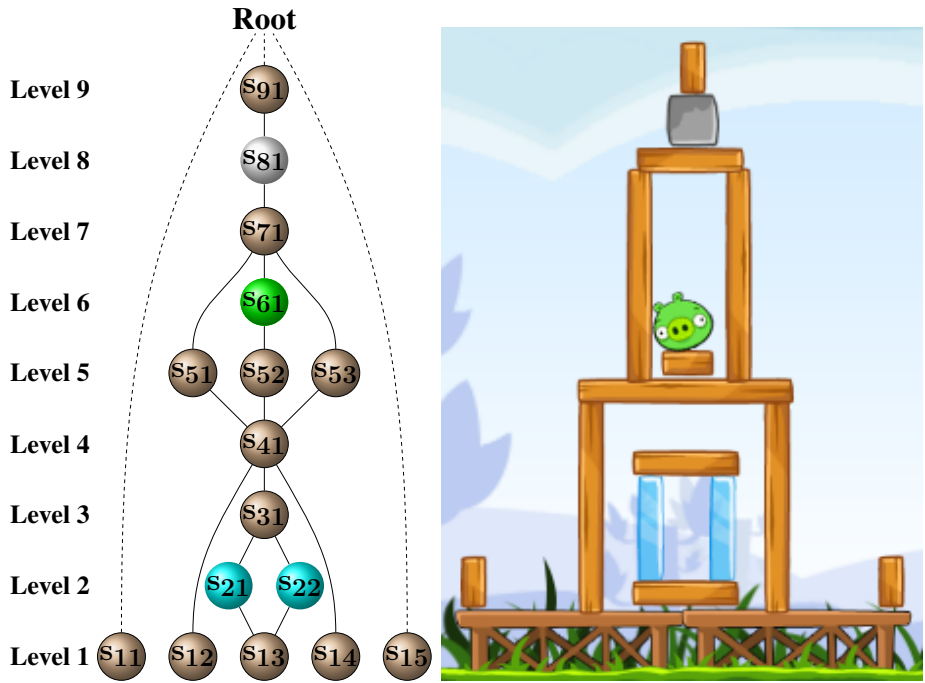

Our state space representation follows a \emph{tree-like structure} of
the game scene using spatial abstractions and topological informations. In particular, we construct 
a tree where each node represents a union of adjacent objects of the same material. 
This is done in an hierarchical fashion (\emph{bottom-up}). The root node is considered as a virtual node that communicates with \emph{orphans} nodes, i.e. nodes which do not have any other object above, see for example nodes: $s_{11}, s_{15}, s_{91}$ in Fig.~\ref{fig1}.

Then, we evaluate each node ($s$) of the tree using three quantities:
\begin{itemize}
\item $x_1(s)$: {\bf Personal weight} calculated as the product of the area
$Area(s)$ 
of the object with a coefficient $c_s$ which is related to the type of the objects,
i.e. $x_1(s) = Area(s) \times c_s$.
All types of object have the same value for this coefficient, $c_s=1$,
except for the types of Pig (P) and TNT (T) which have a larger value of $c_s=10$. 

\item $x_2(s)$: {\bf Parents cumulative weight} calculated by the sum of personal 
weights of the node's parents, ${\cal P} (s)$, in the tree, 
i.e. $x_2(s) = \sum_{s' \in {\cal P} (s)} x_1(s')$.

\item $x_3(s)$: {\bf Distance} (in pixels) to the nearest pig, normalized to $[0, 1]$.
This is made dividing the original distance by 100, where we assumed that 100 pixels 
is the maximum distance in the scene among objects and pigs.
\end{itemize}

The above strategy introduces an appropriate and powerful feature space for all the possible targets.
An example of this mechanism is presented in Fig.~\ref{fig1} where 
illustrates the produced tree structure for the scene of the first level
of the game's episode. In addition, Table~\ref{table1} gives the features of 
the constructed tree nodes. 

\begin{table}[h!]
\begin{center}
\label{table1}
\resizebox{0.7\textwidth}{!}{
\begin{tabular}{|*{7}{c|}}
\hline
 & \multicolumn{6}{c|}{Features} \\ \cline{2-7}
Nodes &  Level & Type & Feasible & \parbox{1cm}{\centering Personal \\ Weight ($x_1$)} & \parbox{1cm}{\centering Above \\ Weight ($x_2$)} &  \parbox{1cm}{\centering Distance \\ ($x_3$)}\\ \hline
$s_{11}$ & 1 & Wood & {\bf True}  & 65   & 0    & 0.818 \\ \hline
$s_{12}$ & 1 & Wood & {\bf True} & 312  & 3557 & 0.501 \\ \hline
$s_{13}$ & 1 & Wood & False & 156  & 7656 & 0.660 \\ \hline
$s_{14}$ & 1 & Wood & False & 312  & 3557 & 0.501 \\ \hline
$s_{15}$ & 1 & Wood & False & 65   & 0    & 0.818 \\ \hline 
$s_{21}$ & 2 & Ice  & False & 162  & 3682 & 0.504 \\ \hline
$s_{22}$ & 2 & Ice  & False & 130  & 3682 & 0.504 \\ \hline
$s_{31}$ & 3 & Wood & False & 125  & 3557 & 0.341 \\ \hline
$s_{41}$ & 4 & Wood & False & 318  & 3239 & 0.151 \\ \hline
$s_{51}$ & 5 & Wood & {\bf True} & 318  & 377  &  0.164 \\ \hline
$s_{52}$ & 5 & Wood & False & 72   & 1777 & 0.082 \\ \hline
$s_{53}$ & 5 & Wood & False & 318  & 377  & 0.198 \\ \hline
$s_{61}$ & 6 & Pig     & {\bf True} & 1400 & 377  & 0.170 \\ \hline
$s_{71}$ & 7 & Wood & {\bf True} & 156  & 221  & 0.431 \\ \hline
$s_{81}$ & 8 & Stone & {\bf True}& 156  & 65   & 0.521 \\ \hline
$s_{91}$ & 9 & Wood  & {\bf True}& 65   & 0    & 0.651 \\ \hline  
\end{tabular}}
\caption{The feature vectors along with the feasible and type labels for the $16$ tree nodes of Fig. \ref{fig1}.}
\end{center}
\end{table}

\subsection{Feasibility examination}

The next step to our approach is to examine each 
node in terms of its possibility to be reached. Infeasible 
situations could be happened as the bounding boxes of 
objects in the scene may not be able to perfectly fit these
structures and they often have irregular non-convex shapes. 
In addition, it is possible some obstacles and stable structures 
such as mountains, to be inserted between the slingshot and the
target. Therefore, an examination step is required at each node so as to ensure that
the corresponding trajectories can reach the target.

\begin{figure}[h!]
  \centering
  \begin{tabular}[c]{c}
  	\includegraphics[scale=0.5]{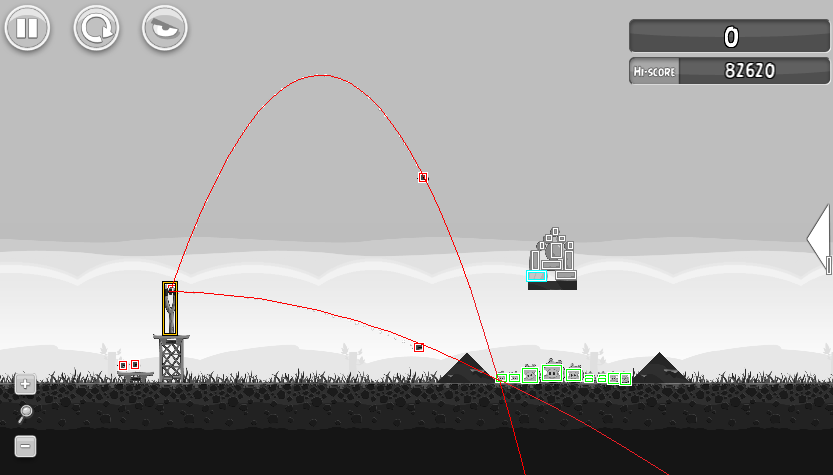} \\
 	(a) \\
	\includegraphics[scale=0.5]{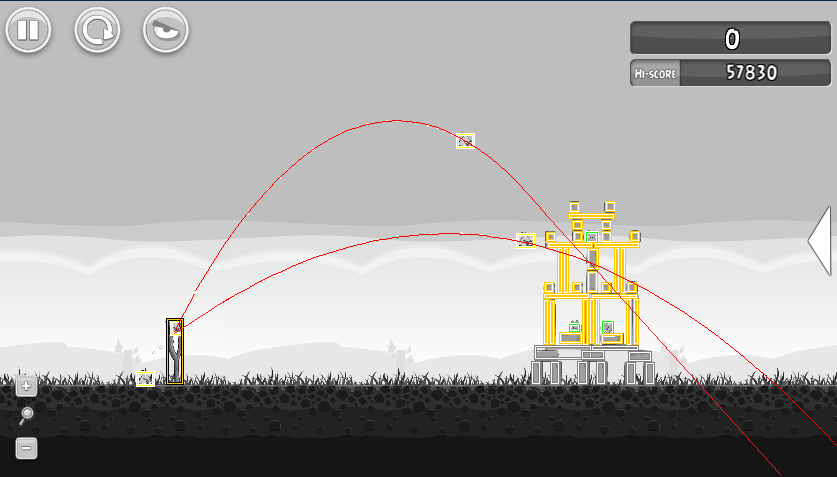} \\
	(b)
  \end{tabular}
\caption{Tree's node feasibility examination. (a) Represents a feasible node (pig) as it is reachable by at least one trajectory. The direct shot is infeasible due to the fact that a mountain is interposed between the slingshot and the target. (b) An infeasible node (wood) is represented as it is not directly reachable due to the tree structure.} 
\label{fig:feasible}
\end{figure}

It must be noted that two different trajectories are calculated, a direct shot (angle $<= 45^{\circ}$) and a high arching shot (angle $> 45^{\circ}$). Both of them are examined in order to estimate the tree's nodes feasibility, see Fig. \ref{fig:feasible}. If there is at least one shot that could reach that node (target) directly, we label it as \emph{feasible} (Fig. \ref{fig:feasible}(a)), otherwise  the tree's node is labeled as \emph{infeasible} (Fig. \ref{fig:feasible}(b)). In the case where both trajectories are accepted, priority is given on the direct shot due to its effectiveness. 
Finally, in the case of the white bird a node is considered as feasible 
if it can be reached by bird's \emph{egg} (Fig. \ref{fig:whitebird}), as opposed to
the other types of birds. 

\subsection{Ensemble of linear regression models}

In our approach we convert the problem of selecting an object for shooting
into an ensemble regression framework. We consider the reward values as the real target 
values $t_n$ of samples (feature vectors) $\bm{x}_n$ which are observed sequentially. They correspond to noisy 
measurements of the output of an $M$-order linear regression model together with an additive 
noise term $\epsilon_n$:
\begin{equation*}
t_n = \sum_{i=1}^M w_i \phi_i(x_n)  + \epsilon_n = \bm{w}^{\top} \bm{\phi}(\bm{x}_n) + \epsilon_n 
\mbox{ ,}
\end{equation*}
where $\bm{w}=(w_1, \ldots, w_M)^{\top}$ is the vector with the $M$ unknown regression parameters.
The above equation represents the reward as a linearly weighted sum of $M$
fixed basis functions denoted by $\bm{\phi}(\bm{x}) = (\phi_1(\bm{x}), \phi_2(\bm{x}), \ldots, \phi_M(\bm{x}))^{\top}$.
The error term $\epsilon$ is assumed to be zero mean Gaussian with variance $1/\beta$, 
i.e. $\epsilon \sim {\cal N}(0,\beta^{-1})$.

Specifically, we have considered Gaussian kernels as basis functions 
following the next procedure: At first we have gathered a number 
of data (feature vectors) from different scenes of the game. 
Then, we performed an agglomerative hierarchical clustering procedure to them, 
where we have applied the standardized Euclidean distance for the merging procedure. 
Finally, we have selected a number $M$ of clusters, where we calculated their mean 
$m_{ik}$ and variance $s^2_{ik}$ for any feature ($k=\{1,2,3\}$). 
Therefore, kernel functions have the following form: 
\begin{equation*}
\phi_i(\bm{x}) = \exp \left( - \sum_{k=1}^3 \frac{(x_k-m_{ik})^2 }{2s^2_{ik}} \right)
\mbox{ .}
\end{equation*}
It must be noted that the number of clusters was not so crucial for the performance of the method. 
During our experimental study we have found that a number of $M=150$ clusters was adequate.

Consider a sequence of observations (input vectors) $\{\bm{x}_k\}_{k=1}^n$ along with the
corresponding targets $t_{1:n}=\{t_k\}_{k=1}^n$.
Therefore, given the set of regression parameters ${\bm{w}, \beta}$ we can model the conditional
probability density of the targets $t_{1:n}$ with the normal distribution, i.e.
\begin{equation*}
p(t_{1:n} | \bm{w}, \beta) = {\cal N} (t_{1:n} | \Phi_n \bm{w}, \beta^{-1}I_n)
\mbox{ ,}
\end{equation*}
where matrix $\Phi_n=[\bm{\phi}(\bm{x}_1), \bm{\phi}(\bm{x}_2), \ldots, \bm{\phi}(\bm{x}_n)]^{\top}$ is called the
design matrix of size $n \times n$ and $I_n$ is the identity matrix of order $n$.

An important issue, when using a regression model is how to
define its order $M$, since models of small order may
lead to underfitting, while large values of $M$ may lead to
overfitting. One approach to tackle this problem is through the Bayesian
regularization method that has been
successfully employed at \cite{Tipping01, Bishop06}. 
According to this scheme, a zero-mean (spherical) Gaussian prior distribution
over weights $w$ is considered:
\begin{equation*}
p(\bm{w} | \alpha) = {\cal N}(\bm{w} | \bm{0}, a^{-1} I) \mbox{,}
\end{equation*}
where the hyperparameter $\alpha$ is the common inverse variance of all weights
and $I$ is the identity matrix.
In this direction we can obtain the posterior distribution over the weights $\bm{w}$,
which is also Gaussian, as:
\begin{equation*}
p(\bm{w}|t_{1:n}, \alpha, \beta) = {\cal N} (\bm{w} | \bm{\mu}_n, \Sigma_n)
\mbox{ ,}
\end{equation*}
where its mean and covariance are given by
\begin{equation*}
\bm{\mu}_n = \beta \Sigma_n \Phi_n^{\top} t_{1:n}  \ \mbox{ ,} \  \ 
\Sigma_n = (\beta \Phi_n^{\top} \Phi_n + aI)^{-1}  \mbox{.}
\end{equation*}

Then, when examining a test point (node) $\bm{x}_*$ we can calculate the prediction and
obtain its corresponding target $t_*$ according to the predictive distribution.
In the Bayesian framework, this is based on the posterior distribution
over the weights, 
\begin{equation*}
p(t_* |t_{1:n}, \alpha, \beta) = {\cal N} (t_* | \bm{\mu}_n^{\top} \bm{\phi}(\bm{x}_*), \beta_*)
\mbox{ ,}
\end{equation*}
where 
\begin{equation*}
\frac{1}{\beta_*} = \frac{1}{\beta} + \bm{\phi}(\bm{x}_*)^{\top} \Sigma_n \bm{\phi}(\bm{x}_*) \mbox{.}
\end{equation*}

Our framework follows an ensemble approach in the sense that
we have a separated regression model for each pair of material object and bird type. 
Totally, there are $7 \times 5=35$ different 
parametric linear regression models, each one has its own set of regression parameters 
$\theta_j = \{ \bm{w}_j, \beta_j \}$. Thus, every time we select a regressor for estimating the 
expected reward per each possible target (node).

In our approach, we have translated the selection mechanism into
a multi-armed bandit problem which offers a trade-off between 
exploration and exploitation during learning. In particular,
we have applied the Upper Confidence Bound (UCB) algorithm \cite{Auer02} for
choosing the next arm (bird-material type regressor) to play.
The selection mechanism is restricted only to the feasible nodes of the current tree. 
According to the UCB, each arm maintains the number of times (frequency) that has been played,
denoted by $n_{f(q)}$, where $f(q)$ corresponds to the type of the regression model 
for the specific node $q$ and the bird type used. The algorithm greedily picks the 
\emph{arm} $f(j^*)$ as follows:
\begin{equation*}
j^* = \argmax_q \left\{ \left(\bm{\mu}^{f(q)}_{n_{f(q)}}\right)^{\top} \bm{\phi}(\bm{x}_q) + C \sqrt{\frac{2 \ln N}{n_{f(q)}}} \right\}
\mbox{ ,}
\end{equation*} 
where $N$ is the total number of plays so far, $\bm{x}_q$ is the feature vector of a node 
and $\bm{\mu}^{f(q)}_{n_{f(q)}}$ is the current estimation of the regression coefficients that corresponds
to the ensemble of the specific bird-material type pair. Finally, $C$ is a constant of the UCB decision making process (during our experiments we have used $C = 3000$).

\subsection{Tap Timing}

\begin{figure}[h!]
  \centering
  \includegraphics[scale=0.5]{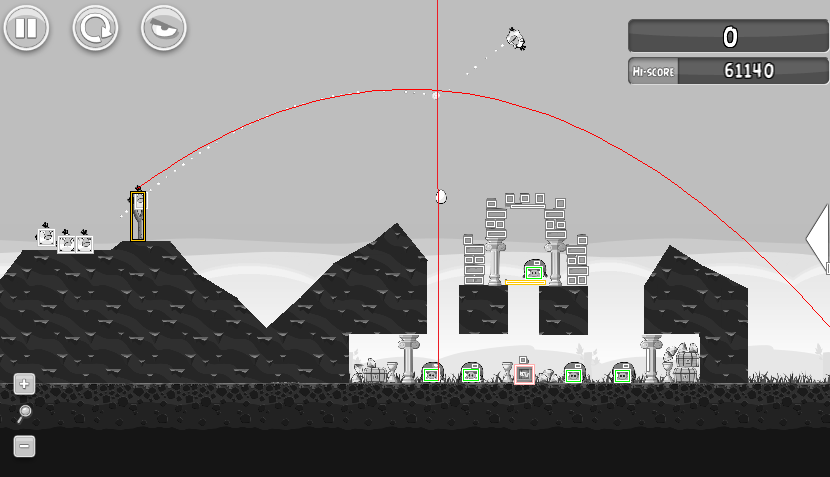}
  \caption{Tap timing procedure for the white bird.} 
\label{fig:whitebird}
\end{figure}

After selecting the target among the feasible nodes of tree, the tap timing procedure is then executed. 
Using the trajectory planner component of the game playing
framework the corresponding tap time is calculated and a tap
is performed right before the estimated collision point.
In our approach the tap time strategy depends on the type of birds used:
\begin{itemize}
\item For the red and black birds (\emph{Bomb} birds are the most powerful among the birds) no tapping is performed.
\item Blue birds (\emph{the Blues}) split into a set of three similar birds when the player taps the screen. The agent performs a tap in an interval between the $65\%$ and $80\%$ of the trajectory from the slingshot to the first collision object.
\item Yellow birds (\emph{Chuck}) accelerate upon tapping which performed  between $90\%$ and $95\%$ of the trajectory in the case of high-arching shots (angle $ > 45^{\circ}$). In the case of direct shots (angle $<= 45^{\circ}$), tap time has been selected randomly between $85\%$ and $90\%$ of the trajectory.
\item White birds (\emph{Matilda}) drop \emph{eggs} in the target below them. In this case tapping is executed when the bird lies above the target (see, Fig.~\ref{fig:whitebird}). As experiments have shown, this strategy is very efficient for handling this specific type of birds. 
\end{itemize}

\subsection{Online learning of model parameters}

The final step of the proposed scheme is the learning procedure.
Due to the sequential nature of data, we have followed a recursive
estimation framework for updating the regression model parameters \cite{Bishop06}.
This can be considered as an online learning solution to
the Bayesian learning problem, where the information on the parameters is
updated in an online manner using new pieces of information (rewards)
as they arrive.
The underlying idea is that at each measurement
we treat the posterior distribution of previous time step as the prior for the
current time step.

Suppose that we have selected a regressor, $k \defn f(j^*) $, for making the prediction
upon an object that has a feature vector $\bm{x}_{n_{k}+1}$. 
After the tapping procedure we receive a reward $t_{n_{k}+1}$.
The recursive estimated solution is obtained by using the posterior
distribution conditioned on the previous $n_k$ measurements $t_{1:n_k}$:
\begin{equation*}
p(\bm{w}_{k}|t_{1:n_{k}})=
{\cal N}(\bm{w}_{k}|\bm{\mu}^{k}_{n_{k}},\Sigma^{k}_{n_{k}})\mbox{.}
\end{equation*}
The new received observation (reward) $t_{n_{k}+1}$ follows the distribution
$p(t_{n_{k}+1}|\bm{w}_{k})=
{\cal N}(t_{n_{k}+1}|\bm{w}_{k}^T \bm{\phi}(\bm{x}_{n_k+1}),\beta_{k})$.
Thus, we can obtain the posterior distribution of weights as: 
\begin{eqnarray*}
p(\bm{w}_{k}|t_{1:n_{k}+1}) & = & p(t_{n_{k}+1}|\bm{w}_{k}) p(\bm{w}_{k}|
                                         t_{1:n_{k}}) \\ 
                    & = & {\cal N} (\bm{w}_{k}|\bm{\mu}^{k}_{n_{k}+1},\Sigma^{k}_{n_{k}+1}) 
                    \mbox{ ,}
\end{eqnarray*}
where the Gaussian parameters can be written in a recursive fashion as:
\begin{eqnarray*}
\Sigma^{k}_{n_{k}+1} & = & \big[ (\Sigma^{k}_{n_{k}})^{-1} + 
      \beta_{k} \bm{\phi}(\bm{x}_{n_{k}+1})^T \bm{\phi}(\bm{x}_{n_{k}+1})\big ]^{-1}, \\
\bm{\mu}^{k}_{n_{k}+1} & = & \Sigma^{k}_{n_{k}+1} \big [ \beta_{k} 
      \bm{\phi}^T(\bm{x}_{n_{k}+1}) t_{n_{k}+1}  +  
                     (\Sigma^{k}_{n_{k}})^{-1} \bm{\mu}^{k}_{n_{k}} \big ].
\end{eqnarray*}

The above equations constitute a recursive estimation procedure for the regression
model parameters.
In the beginning of the estimation (i.e. step $0$) all the information we have about
the model parameters $\bm{w}_k$, is the prior distribution $p(\bm{w}_k)$ which is 
assumed to be zero mean Gaussian ($\bm{\mu}^k_0=\bm{0}$) with spherical
covariance matrix ($\Sigma^{k}_0 = a^{-1} I$).
A last note is that, the sequential nature of estimation allows us to 
monitor the effect of learning progress to parameters.

\section{EXPERIMENTAL RESULTS}

A series of experiments has been conducted in an attempt to analyze 
the performance of the proposed agent (\textbf{AngyBER}) in the Angry birds domain. 
Due to the low complexity of the general framework of our agent, 
the experiments took place in a conventional PC\footnote{Intel 
Core 2 Quad (2.66GHz) CPU with 4GiB RAM}.  

Our analysis was concentrated mainly on the first 21 levels of the freely available 
`Poached Eggs' episode of Angry Birds. 
During the learning phase of the AngryBER agent, 
a complete pass of the previously mentioned episode was executed
more than once (in our study we have passed the episode 10 times). 
For comparison purposes, 
we have used the default \emph{naive} agent, as well as the published results of
the participant teams of the last IJCAI 2013 Angry Birds competition, since they
are provided by the the organizers of the competition\footnote{https://aibirds.org/benchmarks.html}. 
During testing, we have tried to follow the instructions mentioned in the competition rules, 
by setting a time limit of 3 minutes per level on average, that is, a total time of 63 minutes 
for the 21 levels. It must be noticed that our agent requires approximately 
forty (40) minutes for a successfully episode completion.

The depicted results are presented in Table \ref{ResultsTable} 
that gives statistics about the performance of the AngryBER agent, 
i.e. mean values and stds of the score reached per game level.
Note that (after learning) we have made 10 independent runs of the episode.
More specifically,  mean and standard deviation of the score received per level, 
averaged over 10 runs.  
Furthermore, the maximum and minimum received score per level is also given. 

The first remark that stems from our empirical evaluation is that our AngryBER agent 
achieves to pass every level with success at each run. 
Apart from a small fraction, AngryBER achieves to gain quite large scores in the majority of levels. 
That is interesting to be noted is the fact that our agent 
obtains the highest score in seven (7) levels as highlighted in Table~\ref{ResultsTable}, 
comparing with the results of all other agents of the last year's competition. 
At the same time, the mean accumulative score received per episode 
is approximately equal to the highest total score achieved 
among all the other agents. 

Another impressive characteristic of the proposed scheme 
is its ability to speed-up learning process and to discover near optimal policies
quite fast. We believe that this is happened due to the tree structure representation 
in combination with the ensemble strategy. This allows AngyBER agent to be
specialized at each possible  pair material-bird type, recognizing the special bird's behavior on specific materials.
Last but not least, it must be noted that we have conducted a number preliminary
experiments on Levels 22-42, where the results were similar making the generalization 
ability of our approach more evident.

\begin{table}
\begin{center}
\resizebox{1\textwidth}{!}{
\begin{tabular}[c]{|*{6}{c|}}
\hline
\textbf{Level}	 & \multicolumn{3}{c|}{\textbf{AngryBER Agent}} 
 & \textbf{Naive Agent} & \textbf{High scores of IJCAI 2013}\\ \cline{2-4}
 & \textbf{Mean Scores} & \textbf{Max Scores} & \textbf{Min Scores} & 
 & \textbf{Angry Birds Competition} \\ \hline
$1$  & 28740  $\pm$ 165.6 & 28940 & 28400 & 29510 & \textbf{31210} \\ \hline
$2$  & 51370 $\pm$ 2875.1&52360& 43190 & 52230 &  \textbf{60400}\\ \hline
$3$  & 41917 $\pm$ 9.5 & 41920 & 41890 & 40620 & \textbf{42240}	\\ \hline
$4$  & 27049 $\pm$ 3485.6 & 29110& 20350 & 20680	& \textbf{36770} \\ \hline
$5$  & 65483 $\pm$ 2272.9 & \textbf{69800}& 63350 & 55160 & 65850	\\ \hline
$6$  & 33961 $\pm$ 2860.0 & 35200& 26020 & 16070	& \textbf{36180} \\ \hline
$7$  & 26449 $\pm $ 7767.8 & 45650& 20430 & 21590	& \textbf{49120} \\ \hline
$8$  & 53191 $\pm$ 8782.2 & 57110& 28240 & 25730	& \textbf{57780} \\ \hline
$9$  & 36053 $\pm$ 7392.9 & \textbf{52320} & 24410 & 35490 & 51480	\\ \hline
$10$ & 50547 $\pm$ 11221.9 & 65560& 37980 & 32600 & \textbf{68740} \\ \hline
$11$ 	& 55211 $\pm$ 7756.4 & \textbf{60030}& 33490 & 46760 & 59070	 \\ \hline
$12$ & 50151 $\pm$ 5502.5 & 54800& 36530 & 54070 & \textbf{58600} \\ \hline
$13$ & 43945 $\pm$ 7214.3 & \textbf{50920}& 25200 & 49470 & 50360 \\ \hline
$14$ & 70181 $\pm$ 7176.1 & \textbf{79330}& 56620 & 50590 & 65640 \\ \hline
$15$ & 43185 $\pm$ 3998.4 & 51620& 38460 & 46430 & \textbf{55300} \\ \hline
$16$ & 60430 $\pm$ 3295.1 & 63650& 53680 & 55210 & \textbf{66550} \\ \hline
$17$ & 48242 $\pm$ 3745.8 & 52050& 39760 & 48140 & \textbf{54750} \\ \hline
$18$ & 42975 $\pm$ 3145.8 & 48480& 40210 & 49430 & \textbf{54500} \\ \hline
$19$ & 30622 $\pm$ 4533.6 & \textbf{39110} & 21130 & 37920 & 38460	 \\ \hline
$20$ & 45523 $\pm$ 5643.8&	54370          & 38870&36790 & \textbf{56050} \\ \hline
$21$ & 66012 $\pm$ 5911.5 & \textbf{78100} &58760 & 54240 & 75870	\\  \hhline{|=|=|=|=|=|=|}
\textbf{Total} & 971237 $\pm$14647 & 991370 & 943250&858730 & \textbf{1134920} \\ \hline
\end{tabular}}
\label{ResultsTable}
\caption{Performance statistics of the proposed agent in the first 21 levels of the `Poached Eggs' episode}
\end{center}
\end{table}

\section{CONCLUSIONS AND FUTURE WORK}

In this work, we presented an advanced intelligent agent 
for playing the Angry Birds game based on an ensemble of 
regression models. 
The key aspect of the proposed method lies on the efficient 
representation of state space as a tree structure and the 
exploitation of its superior modeling capabilities to establish 
a rich feature space. An ensemble scheme of Bayesian regression
models is then presented, where different bird-material type of regressors 
over the tree are combined and act as ensemble members in a competitive fashion.
The best prediction is then selected for
the decision making process. Learning in the proposed scheme
is achieved in terms of an online estimation framework.
Initial experiments on several game levels 
demonstrated the ability of the proposed methodology to achieve
improved performance and robustness compared to other
approaches on the Angry Birds domain.

We are planning to study the performance of the proposed
methodology to other game levels and test its generalization
capabilities more systematically.
Since the tree structure is very effective and convenient,  
another future research direction is to examine the possibility 
to enrich the feature space with other alternative topological features
which can be extracted for the proposed lattice structure.
A general drawback in the regression analysis is how to
define the proper number of basis functions.
Sparse Bayesian regression offers a solution to the model
selection problem by 
introducing sparse priors on the model parameters
\cite{Tipping01}, \cite{Seeger08}, \cite{Blekas14}.
During training, the coefficients
that are not significant are vanished due to the prior, thus
only a few coefficients are retained in the model which are
considered significant for the particular training data.
This constitutes a possible direction for our future work
that may improve further the proposed methodology.  


\bibliography{angrybirds2014}

\end{document}